%% file: aistats_jrtsub.tex
\documentclass[twoside]{article} 
\usepackage{aistats2017}
\usepackage{etex}
\usepackage{natbib}

\usepackage[utf8]{inputenc} 
\usepackage[T1]{fontenc}    
\usepackage{hyperref}       
\usepackage{url}            
\usepackage{booktabs}       
\usepackage{amsfonts}       
\usepackage{nicefrac}       
\usepackage{microtype}      

\usepackage{graphicx}
\usepackage{subfigure}
\usepackage{caption}

\usepackage{tikz}
\usetikzlibrary{bayesnet, shapes, arrows, positioning, calc, patterns, shadows, external}
\usepackage{verbatim}
\usepackage{lmodern}
\usepackage{scrextend}
\usepackage{relsize}
\usepackage{pgfplots}
\usepgfplotslibrary{groupplots}
\usepackage{pgfplotstable}

\usepackage{amsmath}
\usepackage{amssymb}
\usepackage{bbm}
\usepackage{bm}

\usepackage{booktabs}
\usepackage{colortbl}
\usepackage{multirow}

\usepackage{standalone}

\usepackage{algorithm}
\usepackage{algorithmic}

\usepackage{hyperref}

\usepackage{array}
\newcolumntype{L}[1]{>{\raggedright\let\newline\\\arraybackslash\hspace{0pt}}m{#1}}
\newcolumntype{C}[1]{>{\centering\let\newline\\\arraybackslash\hspace{0pt}}m{#1}}
\newcolumntype{R}[1]{>{\raggedleft\let\newline\\\arraybackslash\hspace{0pt}}m{#1}}

\usepackage[normalem]{ulem}

\begin{document}

%

%

\twocolumn[

\aistatstitle{Infinite Mixture Model of Markov Chains}

\aistatsauthor{ Jan Reubold \And Thorsten Strufe \And Ulf Brefeld }

\aistatsaddress{ TU Dresden \And TU Dresden \And Leuphana University } ]

\begin{abstract}
We propose a Bayesian nonparametric mixture model for prediction- and information extraction tasks with an efficient inference scheme. It models categorical-valued time series that exhibit dynamics from multiple underlying patterns (e.g. user behavior traces). 
We simplify the idea of capturing these patterns by hierarchical hidden Markov models (HHMMs) - and extend the existing approaches by the additional representation of structural information.
Our empirical results are based on both synthetic- and real world data. They indicate that the results are easily interpretable, and that the model excels at segmentation and prediction performance: it successfully identifies the generating patterns and can be used for effective prediction of future observations.
\end{abstract}

\input{introduction}
\input{relatedwork}

\input{contribution}

\input{experiments}

\input{conclusion}

{\small
	\bibliography{aistats_jrtsub.bbl}}
\bibliographystyle{plainnat}


\end{document}

%% file: introduction.tex
\section{Introduction} \label{sec:introduction}

Assume that the behavior of users follows intentions, for example in the context of web interaction they may want to look for information on a specific topic or check their e-mails. In order to fulfill an intention, they must complete a number of actions, such as requesting a certain web-page, often in a certain order. If we assume that a similar sequence of actions belongs to the same or a similar intention, we should be able to recognize an intention given a sequence of actions. Furthermore, given an entire data set of such sequences we should be able to identify the intentions themselves by recognizing reoccurring patterns within the data. Generalizing from this idea, one can think of a two-level hierarchy of dynamics. One level representing the sequence of intentions exhibiting so-called high-level dynamics, and one level that represents the sequence of actions performed while fulfilling a specific intention displaying so-called low-level dynamics.  


Data that exhibits these complex dynamics following different patterns (intentions) can be observed in various domains.
These patterns are commonly referred to as super states \citep{johnson2013bayesian}.
In categorical-valued time-series -- series of discrete values where the only known relation between different values is the temporal relation -- these super states are observed as sub-sequences, called segments. Each time-series can be generated by multiple underlying super states. 
Therefore, consecutive observations within a segment possess low-level dynamics while transitions between super states, meaning transitions between segments, exhibit so-called high-level dynamics. 

Modeling such data, with tasks such as identifying the number of super states and their dynamics within a dataset, is a challenging problem. The models need to be very flexible and, thus, get extremely complex very quickly:
Bayesian nonparametric models successfully capture data exhibiting complex low-level dynamics \citep{fox11, beal12}. The general idea is, again, to identify the underlying super states by grouping similar segments.
Approaches that aim at grasping dynamics on different levels struggle with either their efficiency \citep{hhmm} or flexibility.
Nonetheless, such models are crucial to capture natural processes that possess both low- and high-level dynamics, like navigation strategies of users searching for information on the Web \citep{west12} or on Facebook \citep{puscher}, human activities of daily living \citep{duong05}, natural language \citep{lee13}, or motion recognition \citep{heller09}.


The goal of this paper is to develop an approach for the segmentation of categorical-valued time-series data that can be used for prediction- and information extraction tasks. Regarding the model, our requirements are as follows: (i) the algorithm should perform a multi-level analysis, covering at least two levels of the dynamics (e.g. number of intentions and their manifestations), (ii) the number of super states should be unbounded (e.g. one cannot set a bound on the number of intentions), (iii) focus on categorical-valued time-series data (sequences of arbitrary length), (iv) possess some predictive capabilities, and (v) yield results that are easy to interpret. 
The first three requirements relate to the segmentation task, the last two represent equally important requirements for user understanding.

Requirement (ii) suggests using a Bayesian nonparametric treatment. Markov chains (MCs) address (iii) and guarantee a certain amount of predictive power (iv) as well as well interpretable results (v) and a simple inference scheme. 
Additionally, combining both concepts allows us to perform a two-level analysis of the dynamics of the data (ii).
In this paper, we hence propose a Bayesian nonparametric mixture model where each mixture component is represented by a MC. Therefore, the model learns two-level dynamics in an unsupervised fashion and represents each identified super state, encoding (relatively) stable low-level dynamics, by a MC.

The main goal of our research is to enhance both, the prediction of future behavior and the understanding of the dynamics in the context of categorical-valued behavioral data by means of segmentation. 
Therefore, we evaluate the segmentation performance of our model against synthetic data, to understand its effectiveness and test it for extreme cases. 
Further, we apply our model to a novel task of user understanding, where we segment behavior traces of users on Facebook to understand their behavior and predict their next moves. 
Our empirical findings indicate that our model successfully identifies underlying patterns and can effectively be turned into a predictor for future observations.

%% file: relatedwork.tex
\section{Related Work} \label{sec:rw}

Two models that can naturally capture dynamics caused by multiple underlying super states are the standard- and the infinite hierarchical hidden Markov models ([i]HHMM) \citep{hhmm, murphy02, heller09}. Each hierarchy of a [i]HHMM is a separate hidden Markov model (HMM) with all observations situated in the leaves, called production states. Where the HHMM requires an a-priori fixed number of levels for its hierarchy, the iHHMM allows for a potentially unbounded number that can grow with data. 
Due to the unbounded depth of the hierarchy of HMMs, these models are highly flexible. Nonetheless, they are rather simple with respect to the structural information used. Each hierarchy consists of HMMs without any further structural information incorporated. To the best of our knowledge, there exists no extension that incorporate additional, structural information due to the complicated and expensive inference in these models. In the classical model the inference scheme rendered the (i)HHMM inapplicable to real-world problems \citep{hhmm, heller09}, until \citet{wakabayashi2012forward} developed a more efficient one.
Due to studies that suggest that two-level analyses of dynamics are sufficient in many real-world applications \citep{oliver04, nguyen05, xie03}, related work simplifies the iHHMM by restricting the depth of the hierarchy while integrating additional structural information.

\citet{stepleton09} propose a model where the infinite HMM \citep{beal01} (iHMM) is combined with a block-diagonal prior. The model assumes that the transition matrix of the iHMM is comprised of a nearly block-diagonal structure. It groups subsets of hidden states into blocks, generating an unbounded number of blocks. By modifying the Dirichlet process prior over the transitions, the model increases the transition probability of states within a block. Each block can be interpreted as a super state. However, the model cannot handle super states with overlapping categorical-valued state spaces.
A similar idea, a bias towards self-transitions within a mixture component of the hierarchical Dirichlet process - HMM \citep{teh06} (HDP-HMM), is an essential part of the sticky HDP-HMM \citet{fox11} propose. In similarity to block-diagonal iHMM, successive hidden states in this model favor to belong to the same state. Further, by augmenting the hidden states with an additional layer of states, the sticky HDP-HMM allows to treat the conditional distribution of observations given the states nonparametrically. While the model is able to partition sequences into segments, it is not applicable to categorical-valued time-series, whose values only stand in temporal relations to each other. Furthermore, the model cannot capture any dynamics within a super state.

Studies by \citet{johnson14} and \citet{saeedi16} explore the benefits of incorporating an explicit state-duration distribution instead of defining some bias towards specific transitions \citep{fox11, stepleton09}. Both approaches are Bayesian nonparametric models that apply a two-level analysis of the dynamics within the data. Whereas the model proposed by \citet{johnson14} learns a distribution expressing the overall duration of a state, the segmented iHMM (siHMM) \citep{saeedi16} models a state-duration distribution which expresses the probability of changing the current state, conditioned on the current observation and hidden state. 
Similar to the sticky HDP-HMM, both models cannot capture the dynamics within a super state.
In general, none of the existing approaches fulfills all requirements and only the (i)HHMM satisfy our requirements for segmentation (i-iii) without further adaptation.

Finally, \citet{cadez2000visualization} propose an finite mixture model of Markov chains (FMMC). While, due to its parametric nature, it is not flexible enough for segmentation, the concept behind this algorithm is similar to ours, i.e. a mixture of Markov chains.

\input{gms}

Our model combines aspects of both concepts, i.e. it incorporates a bias towards self-transitions as well as a natural state-duration model by identifying the distribution over the start- and end states of each super state. It features a simple inference scheme and fulfills the requirements, e.g. the obtained model inherently features prediction tasks. 



%% file: gms.tex
    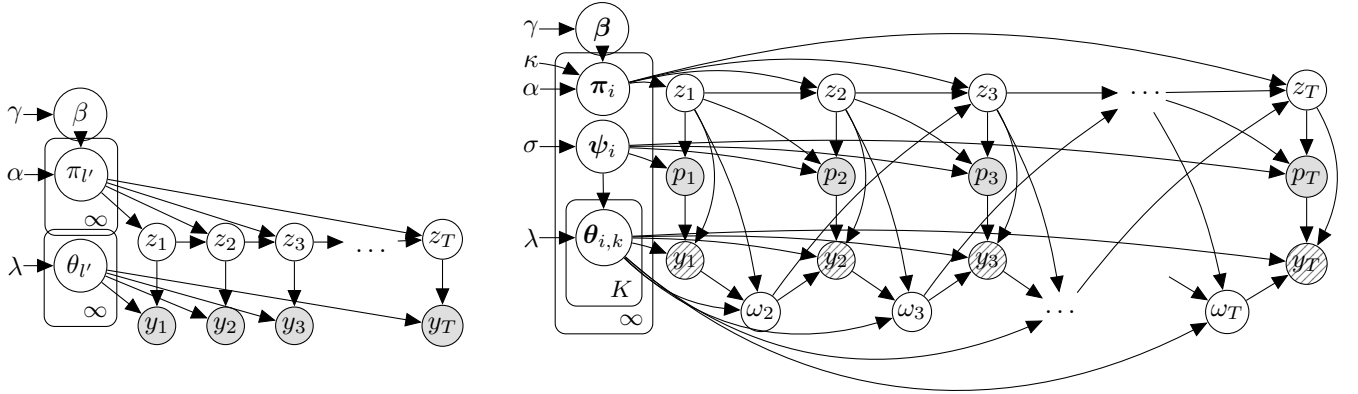
\begin{figure*}[t]
        \begin{minipage}{0.16\textwidth} 
            \begin{tikzpicture}[x=0.4cm,y=0.6cm]


                \node[const]                              (lambda) {$\lambda$} ;
                \node[const , above=of lambda, yshift=0.4cm]  (alpha)  {$\alpha$}  ;
                \node[const , above=of alpha]                 (gamma)  {$\gamma$}  ;
                \node[latent, right=of lambda]              (theta)  {$\theta_{l'}$};
                \node[latent, above=of theta , right=of alpha](pi)     {$\pi_{l'}$}   ; %
                \node[latent, above=of pi    , right=of gamma](beta)   {$\beta$}   ; %
                \node[obs   , right=of theta , yshift=-0.8cm, minimum size=5mm](y1)     {$y_1$};
                \node[obs   , right=of y1, minimum size=5mm]    (y2)      {$y_2$}    ;
                \node[obs   , right=of y2, minimum size=5mm]    (y3)      {$y_3$}    ;
                \node[const , right=of y3, minimum size=5mm,  yshift=1.0cm](dots)   {$\hspace{0.15cm}\dots$};
                \node[obs   , right=of dots, minimum size=5mm, yshift=-1.0cm]  (yT)     {$y_T$}     ;
                \node[latent, above=of y1, minimum size=5mm]    (z1)      {$z_1$}    ;
                \node[latent, above=of y2, minimum size=5mm]    (z2)      {$z_2$}    ;
                \node[latent, above=of y3, minimum size=5mm]    (z3)      {$z_3$}    ;
                \node[const , right=of z3, minimum size=5mm]    (nan)     {$\qquad$} ;
                \node[latent, above=of yT, minimum size=5mm]    (zT)      {$z_T$}    ;

                \edge {alpha}  {pi}    ; %
                \edge {lambda} {theta} ;
                \edge {gamma}  {beta}  ;
                \edge {beta}   {pi}    ;
                \edge {theta}  {y1}    ;
                \edge {theta}  {y2}    ;
                \edge {theta}  {y3}    ;
                \edge {theta}  {yT}    ;
                \edge {pi}     {z1}    ;
                \edge {pi}     {z2}    ;
                \edge {pi}     {z3}    ;
                \edge {pi}     {zT}    ;
                \edge {z1}     {z2}    ;
                \edge {z2}     {z3}    ;
                \edge {z3}     {nan}   ;
                \edge {nan}    {zT}    ;
                \edge {z1}     {y1}    ;
                \edge {z2}     {y2}    ;
                \edge {z3}     {y3}    ;
                \edge {zT}     {yT}    ;

                \plate {plate1} { %
                (theta) %
                } {$\infty$}; %
                \plate {ppi} { %
                (pi) %
                } {$\infty$} ; %

            \end{tikzpicture}
        \end{minipage}
        \hfill
        \begin{minipage}{0.6\textwidth}
            \begin{tikzpicture}[x=0.48cm,y=0.6cm]

                \node[const]                                   (lambda2){$\lambda$};
                \node[const , above=of lambda2, yshift=0.4cm]  (sigma) {$\sigma$};
                \node[const , above=of sigma]                  (alpha2){$\alpha$};
                \node[const , above=of alpha2, yshift=-0.4cm]   (rho)   {$\kappa$};
                \node[const , above=of alpha2]                 (gamma2){$\gamma$};
                \node[latent, right=of lambda2]                (theta2){$\bm{\theta}_{i,k}$};
                \node[latent, right=of sigma]                  (psi)   {$\bm{\psi}_{i}$};
                \node[latent, above=of psi    , right=of alpha2](pi2)  {$\bm{\pi}_{i}$};
                \node[latent, above=of pi2    , right=of gamma2](beta2){$\bm{\beta}$};
                \node[const , below=of theta2 , yshift=0.2cm, xshift=-0.5cm](ph) {};
                \node[const , below=of theta2 , yshift=0.2cm, xshift=0.5cm](ph2) {};
                
                \node[obs, right=of psi, yshift=-0.4cm, minimum size=5mm] (s1)     {$p_1$};
                \node[latent, pattern color=gray, pattern=north east lines, below=of s1 , minimum size=5mm](y12){$y_1$};
                \node[latent, above=of s1, minimum size=5mm] (z12)    {$z_1$};
                \node[latent, right=of s1, yshift=-1.8cm, minimum size=5mm] (w12)    {$\omega_2$};
                
                \node[obs, right=of w12, yshift=1.8cm, minimum size=5mm] (s2)     {$p_2$};
                \node[latent, pattern color=gray, pattern=north east lines, below=of s2, minimum size=5mm] (y22)    {$y_2$};
                \node[latent, above=of s2, minimum size=5mm] (z22)    {$z_2$};
                \node[latent, right=of s2, yshift=-1.8cm, minimum size=5mm] (w22)    {$\omega_3$};
                                
                \node[obs, right=of w22, yshift=1.8cm, minimum size=5mm] (s3)     {$p_3$};
                \node[latent, pattern color=gray, pattern=north east lines, below=of s3, minimum size=5mm] (y32)    {$y_3$};
                \node[latent, above=of s3, minimum size=5mm] (z32)    {$z_3$};
                
                \node[const , right=of z32, xshift=1.0cm, minimum size=5mm](dots2){$\hspace{0.15cm}\dots$};
                \node[const , below=of dots2, minimum size=5mm] (nan2)   {$\qquad$};
                \node[const , below=of nan2, minimum size=5mm] (nan2p)   {$\qquad$};
                \node[const , right=of s3, yshift=-1.8cm, minimum size=5mm] (wnan) {$\dots$};
                
                \node[latent, right=of nan2, yshift=-1.8cm, minimum size=5mm] (wT2)    {$\omega_{T}$};
	            
                \node[obs, right=of wT2, yshift=1.8cm, minimum size=5mm] (sT)     {$p_T$};
                \node[latent, pattern color=gray, pattern=north east lines, below=of sT, minimum size=5mm](yT2)    {$y_T$};
                \node[latent, above=of sT, minimum size=5mm] (zT2)    {$z_T$};

                \edge {alpha2} {pi2}   ; %
                \edge [bend left=10] {rho} {pi2};
                \edge {lambda2}{theta2};
                \edge {gamma2} {beta2} ;
                \edge {sigma}  {psi}   ;
                \edge {beta2}  {pi2}   ;
                \edge {psi} {theta2};
                \edge [bend left= 5] {theta2} {y12}   ;
                \edge [bend left= 5] {theta2} {y22}   ;
                \edge [bend left= 5] {theta2} {y32}   ;
                \edge [bend left= 5] {theta2} {yT2}   ;
                \edge [bend left=15] {pi2}    {z12}   ;
                \edge [bend left=15] {pi2}    {z22}   ;
                \edge [bend left=15] {pi2}    {z32}   ;
                \edge [bend left=15] {pi2}    {zT2}   ;
                \edge [bend left= 5] {psi}    {s1}    ;
                \edge [bend left= 5] {psi}    {s2}    ;
                \edge [bend left= 5] {psi}    {s3}    ;
                \edge [bend left= 5] {psi}    {sT}    ;
                \edge {z12}    {z22}   ;
                \edge {z22}    {z32}   ;
                \edge {z32}    {dots2} ;
                \edge {dots2}  {zT2}   ;
                \edge {z12}    {s1}    ;
                \edge {z22}    {s2}    ;
                \edge {z32}    {s3}    ;
                \edge {zT2}    {sT}    ;
                \edge [bend left=10] {z12}    {s2}    ;
                \edge [bend left=10] {z22}    {s3}    ;
                \edge [bend left=10] {dots2}  {sT}    ;
                \edge {s1}     {y12}    ;
                \edge {s2}     {y22}    ;
                \edge {s3}     {y32}    ;
                \edge {sT}     {yT2}    ;
                \edge[bend left=30] {z12}    {y12}   ;
                \edge[bend left=30] {z22}    {y22}   ;
                \edge[bend left=30] {z32}    {y32}   ;
                \edge[bend left=30] {zT2}    {yT2}   ;
                \edge {y12}    {w12}    ;
                \edge {y22}    {w22}    ;
                \edge {y32}    {wnan}    ;
                \edge {nan2p}  {wT2}    ;
                \edge [bend left=10] {z12}    {w12}    ;
                \edge [bend left=10] {z22}    {w22}    ;
                \edge [bend left=10] {z32}    {wnan}   ;
                \edge [bend left=10] {dots2}  {wT2}    ;
                \edge [bend right=20] {theta2}  {w12}    ;
                \edge [bend right=31] {theta2}  {w22}    ;
                \edge [bend right=33] {theta2}  {wnan}    ;
                \edge [bend right=35] {theta2}  {wT2}    ;
                \edge {w12}    {y22}    ;
                \edge {w22}    {y32}    ;
                \edge {wT2}    {yT2}    ;
                \edge [bend left=10] {w12}    {z32}    ;
                \edge [bend left=10] {w22}    {dots2}  ;
                \edge [bend left=10] {wnan}   {zT2}    ;

                \plate {inner} {
                (theta2)
                } {$K$};
                \plate {ppi2} { %
                (pi2) %
                (psi)
                (theta2)
                (ph)
                (ph2)
                } {$\infty$} ; %
            \end{tikzpicture}
        \end{minipage}
    \caption{(left) Graphical model of an HDP-HMM \citep{teh06}; (right) graphical model of IMMC; $K$ is defined as $|\Sigma| + 1$, where $+1$ represents the auxiliary boundary node of the lossless concatenation; white nodes, white nodes with gray lines, and gray nodes represent hidden states and partly observed states and observed states, respectively.}
    \label{fig:gms}
    \end{figure*}

%% file: contribution.tex
\section{An Infinite Mixture Model of Markov Chains}
\label{sec:con}
In this section we present our main contribution: the infinite mixture model of Markov chains (IMMC). The model applies a two-level analysis to the dynamics of the data. Compared to the HHMM our approach contains a more detailed state transition model for the super states. The augmentation of both the observation- and the latent state layer results in a natural state duration model with state durations based on the structural information of the dynamics within a super state.
Note that while this paper focuses on the intended use for categorical-valued time-series, such as user traces on online platforms, it is not restricted to these.

We now give a more formal description of the IMMC. Let $\Sigma$ denote a finite observation space and $\Sigma^*$ the set of all sequences of possible combinations over $\Sigma$. Then, $y^{(s)}$ denotes a finite sequence of observations from $\Sigma^*$, with $s$ as its index.
To not clutter the notation unnecessarily, we assume to have a set $\mathbf{Y}$ of $S$ sequences with arbitrary length $T_s$ present as a concatenated sequence $\mathbf{y}$ where the sequences from $\mathbf{Y}$ are separated by an auxiliary boundary-symbol $B$. Therefore, the model can handle sequences of arbitrary length.

The model is comprised of three key parts: (i) The underlying sequence of hidden states assigning an observation to a specific super state is modeled by a HDP-HMM (Fig. \ref{fig:gms}), the equivalent to the iHMM; (ii) the prior information that successive hidden states are more likely to originate from the same super state is expressed by a self-transition bias (as in \citet{fox11}); (iii) finally, to capture the MCs, we augment the layer of the observed states to not represent a single observed state, but transitions between successive observed states. The MCs represent the super states and generate successive sub-states which represent the segments within the sequence of observations. The entire graphical model is depicted in Figure \ref{fig:gms} (right).

The HDP-HMM is a HMM combined with a nonparametric prior that is based on a two-level hierarchy of Dirichlet processes (DPs). A DP is a distribution over distributions. A sample from it, DP($\gamma, H$), can be generated by the `stick-breaking process' of \citet{sethuraman94}. Here, $\gamma$ is called the concentration parameter and $H$ denotes the base measure. The `stick-breaking' process simulates repeatedly breaking a portion from the end of a stick apart. Thinking of the stick as the unit interval, repeatedly breaking a portion of it apart generates a partitioning of the interval, resulting in an infinite set of sub-intervals. Given a positive $\gamma$, the process $\mbox{SBP}_1(\gamma)$ is defined as follows:
\begin{equation} \label{eq:sticka}
    \beta'_i \sim \mbox{Beta}(1, \gamma) \quad \beta_i = \beta'_i \prod^{i-1}_{k=1} (1 - \beta'_k)\quad i = 1, 2, \dots \mbox{  },
\end{equation}
where $\mbox{Beta}(\cdot)$ denotes the Beta distribution, $\beta_i'$ is the fraction of the remaining stick to break of, and $\beta_i$ its total length. $\tilde{\theta}_i$ denotes a realization of an i.i.d. draw from the finite measure, $\tilde{\theta}_i \sim H$. A sample from a DP can then be obtained by
\begin{equation} \label{eq:stickb}
    G = \sum_{k=1}^{\infty} \beta_k \delta_{\tilde{\theta}_k}.
\end{equation}

In a hierarchical Dirichlet process (HDP), which consists of a two-level hierarchy of DPs, the realization of one DP $G$ is used as the base measure for all its subordinate DPs, DP($\alpha, G$). Therefore, these DPs represent distributions over distributions over the same categorical, finite space. Instead of applying Equations \ref{eq:sticka}, \ref{eq:stickb} recursively to sample realizations for both the base DP and its subordinates, \citet{teh06} propose an equivalent scheme, that directly takes the sub-intervals $\beta_i$ as inputs for the `stick-breaking process' of the subordinates. The modified process $\mbox{SBP}_2(\alpha, \beta)$ is given by
\begin{equation}
	\begin{aligned} \label{eq:stick2}
	    \pi'_{ji} &\sim \mbox{Beta}\left(\alpha\beta_i, \alpha \left(1 - \sum_{k=1}^{i} \beta_k\right)\right), \\ \pi_{ji} &= \pi'_{ji} \prod_{k=1}^{i-1} \left(1 - \pi'_{jk}\right).
	\end{aligned}
\end{equation}
Thus, Equations \ref{eq:sticka} and \ref{eq:stick2} are sufficient to realize samples from a HDP. By replacing the set of conditional finite mixture models of the HMM with a HDP, we obtain a nonparametric HMM with an unbounded state space.
To address the problem of fast switching between redundant states in the HDP-HMM to avoid slowing mixing rates and a possible decrease in predictive performance \citep{fox11}, we make use of the mechanism \citet{fox11} propose. Therefore, Equation \ref{eq:stick2} is slightly modified to incorporate a bias towards self-transitions of states,
\begin{equation}
    \pi_{j, \cdot} \sim \mbox{SBP}_2\left(\alpha + \kappa, \frac{\alpha\bm{\beta} + \kappa\delta_j}{\alpha + \kappa}\right), 
\end{equation}
where $\kappa > 0$ is the amount added to the $j$th component and $\bm{\beta} \sim \mbox{SBP}_1(\gamma)$.

The algorithm consists of four layers of states, the hidden states $\mathbf{z}$ and $\mathbf{\omega}$, the observed sub-states $\mathbf{p}$, and the partly observed sub-states $\mathbf{y}$. The hidden state $z_t$ represents the active super state at time-step $1 \leq t \leq T$. The hidden state $\omega_t$ is either $0$ or $1$ and signals the continuation or end of a segment, respectively. Finally, the two sub-state layers, $p_t$ and $y_t$, represent the transition from $p_t$, the sub-state of the previous time-step, to the current sub-state $y_t$. The reason for modeling the transition of sub-states is to identify the dynamics within each super state.
State $y_t$ is defined as partly observed, because we assume that information about the end of a segment is missing in the data. Due to the goal of segmentation, this assumption is necessary.

The resulting generative process is then as follows
\begin{equation}
	\begin{aligned}
	    \bm{\beta} &\sim \mbox{SBP}_1(\gamma) 
	    & \hspace{-0.45cm} \upsilon_k &\sim \mbox{Mu}\left(\{\nicefrac{1}{|\Sigma|}\}^{|\Sigma|}\right) \hspace{-0.25cm}\\[-2pt]
	    \pi_{j, \cdot} & \sim \mbox{SBP}_2\left(\alpha + \kappa, \frac{\alpha\bm{\beta} + \kappa\delta_j}{\alpha + \kappa}\right)\hspace{-1.1cm}
	    &&\hspace{0.3cm} 
	    \psi'_{i, \cdot} & \hspace{-4.5cm}\sim \mbox{SBP}_1(\sigma)\hspace{1.1cm} \\[-5pt]
	    \theta_{i, \cdot, \cdot} &\sim \mbox{SBP}_2\left(\lambda, \psi'_{i, \cdot}\right) 
	    & \psi_{i, \cdot} &= \sum_{k=1}^{\infty} \psi'_{i, k} \delta_{\upsilon_k}\\[-2pt]
	    \omega_t &\sim \mbox{Ber}\left(\theta_{z_{t-1}, y_{t-1}, B}\right)\hspace{-0.3cm} & 
	    z_t\hspace{0.35cm} &\hspace{-0.4cm}\left\{
	    \begin{array}{l l}
	    \sim \pi_{z_{t-1}} & \hspace{-0.08cm} \text{, if $w_{t-1} = 0$}\\
	    = z_{t-1} & \hspace{-0.08cm} \text{, otherwise}
	    \end{array} \right.\\
	    p_t\hspace{0.35cm} &\hspace{-0.4cm}\left\{
	    \begin{array}{l l}
	    \sim \psi_{z_t, \cdot} \hspace{0.15cm} \text{, if $w_{t-1} = 0$}\\
	    = B \hspace{0.52cm} \text{, otherwise}
	    \end{array} \right.\hspace{-0.65cm} &
	    y_t \hspace{0.35cm}&\hspace{-0.4cm}\left\{
	    \begin{array}{l l}
	    \sim \theta_{z_t, p_t, \cdot} \hspace{0.1cm} \text{, if $w_{t} = 0$}\\
	    = B \hspace{0.75cm} \text{, otherwise,}
	    \end{array} \right.
	\end{aligned}
\end{equation}
where $\mbox{Mu}(\cdot)$ denotes the Multinomial distribution, $\mbox{Ber}(\cdot)$ the Bernoulli distribution and $\Sigma$ the finite, categorical-valued sub-state space with cardinality $|\Sigma|$.\\
Note, that, due to the interpretation of the observed layers, we do not process any observation twofold, but process each onetime, i.e. once as the starting- and once as the end state of a transition.\\
The resulting graphical model is depicted in Figure \ref{fig:gms}.

\subsection*{A Blocked Gibbs Sampler}
In this section we present a truncated blocked Markov chain Monte Carlo (MCMC) HDP sampling algorithm, similar to the one \citet{fox11} propose, to optimize the parameters of our model.

\citet{fox11} show that a truncated blocked Gibbs sampler allows to jointly sample hidden states and exploit the Markovian structure. The joint mechanism obtains faster mixing rates than for instance a direct assignment sampler.
To sample distributions of theoretically infinite cardinality, we make use of the degree $L$ weak limit approximation \citep{ishwaran2002exact}, where $L$ denotes the maximum cardinality of the approximated distribution. 
It follows, that in practice $L$ needs to exceed the number of true mixture components. Thus, a DP is approximated by a Dirichlet distribution (Dir), with $\mbox{Dir}(\alpha / L, \dots, \alpha / L)$.
Note that this approximation is commonly used for a simple and more efficient computation (see \citep{fox11}). \citet{kurihara} found little to no practical differences to an inference scheme using no truncation.

The prior distributions $\bm{\beta}, \bm{\pi}, \bm{\psi},$ and $ \bm{\theta}$ are initialized by
\begin{equation}
	\begin{aligned} \label{eq:init_prior}
	    \bm{\beta}  &\sim \mbox{Dir}(\gamma / L, \dots, \gamma / L) & \bm{\pi}_i  &\sim \mbox{Dir}(\alpha\bm{\beta} + \kappa\delta_i)\\
	    \bm{\psi}_i &\sim \mbox{Dir}(\sigma / K, \dots, \sigma / K) & \bm{\theta}_{i,k} &\sim \mbox{Dir}(\lambda\bm{\psi}_i),
	\end{aligned}
\end{equation}
where $1\leq i\leq L$, $K=|\Sigma| + 1$, and $1\leq k \leq K$.

To update the prior distributions after each iteration, we have to keep track of state-, as well as sub-state transitions. Therefore, $d_i$ stores the number of sub-states assigned to each super state $i$ and $G_{i,k_1,k_2}$ records the number of transitions within super state $i$, where $k_1$ and $k_2$ represent the row and column of the transition matrix, i.e. the sub-states of the previous and current time-step, respectively. Finally, $n_{i_1, i_2}$ keeps track of the transitions between super states $i_1$ and $i_2$. For each iteration, the auxiliary variables document the assignment step.

\paragraph*{Sampling $z_t$} We obtain a realization of the hidden states $z_t$ by adapting the Baum-Welch algorithm. For the first pass, applying the algorithm backward in time, from the last to the first observation of the input sequence, we obtain the backward probabilities $m_{t, t-1}$:
\begin{equation} \label{eq:compute_m}
    \begin{split}
        m_{T+1, T}(i) &= 1 \\
        m_{t, t-1}(i) &= \left\{
        \begin{array}{l l}
            m_{t+1, t}(i) & \quad \text{if $r_t = B$};\\
            m_{t+1, t}(i) \cdot \beta_i \cdot \theta_{i, p_t, B} & \quad \text{if $y_t = B$};\\
            \Omega_{t,i} & \quad \text{otherwise}.
        \end{array} \right.
    \end{split}
\end{equation}
At the beginning and end of a new sequence ($r_t = B$ and $y_t = B$) the message of the successive time-step is passed backward. In case of the latter it is weighted with the likelihood of seeing the beginning of a segment instantiated by super state $i$ given $p_t$, the observed state of the previous time-step. Within a sequence, the algorithm has to account for both intra- and inter-transitions. Hereby, intra-transitions account for sub-state transition within a super state, and inter-transitions for state transitions between super states. Therefore, $\Omega_{t,i}$ computes the likelihood of an intra-transition,
\begin{equation}
	L^{\mathsf{intra}}_{t,i} = \theta_{i,p_t,y_t} \cdot \psi_{i,p_t},
\end{equation}
as well as the probability of an inter-transition,
\begin{equation}
	L^{\mathsf{inter}}_{t,i} = \left[\beta_i \cdot \psi_{i,B} \cdot \pi_{i,j}\right] \cdot \left[\psi_{i, r_t} \cdot \theta_{i,r_t,B} \cdot \theta_{j,B,y_t}\right].
\end{equation}

Here, the first part, $\beta_i \cdot \psi_{i,B} \cdot \pi_{i,j}$, represents the prior probability of observing an inter-transition from super state $i$ to $j$. The likelihood of the inter-transition is then expressed by $\psi_{i, r_t} \cdot \theta_{i,r_t,B} \cdot \theta_{j,B,y_t}$.

Given both probabilities, $\Omega_{t,i}$ is computed as follows,
\begin{equation}
	\Omega_{t,i} \propto \sum_{j=1}^L{ \left(L^{\mathsf{intra}}_{i,j} \cdot \mathbb{I}(i=j) + L^{\mathsf{inter}}_{i,j}\right) \cdot m_{t+1,t}(j)}.
\end{equation}

\begin{algorithm}
	{\small
		\vspace{0.3cm}
		\textbf{Given} the hyperparameters $\bm{\beta}, \bm{\pi}, \bm{\psi}, \bm{\theta}, \kappa$
		\begin{enumerate}
			\item Initialize prior distributions according to Eq. \ref{eq:init_prior}\vspace{0.2cm}
			
			\hspace{-0.7cm}\textbf{Until} convergence \textbf{do}:
			\item Perform Baum-Welch algorithm Eqs. \ref{eq:compute_m} - \ref{eq:sample_hidden}
			\item During forward steps update auxiliary variables as follows:
			\begin{itemize}
				\item Increment $d_{z_t}$ \hspace{2.27cm}, if $y_t \neq B$
				\item[] \hspace{1.43cm} $G_{z_t, B, y_t} \land n_{z_{t-1}, z_t}$
				\item[] \hspace{1.43cm} $\land$ \hspace{0.05cm}$G_{z_{t-1}, p_t, B}$\hspace{0.98cm}, if $\omega_t = 1$
				\item[] \hspace{1.43cm} $G_{z_t, r_t, y_t}$ \hspace{1.52cm}, otherwise
			\end{itemize}
			\item Compute posterior distributions according to Eq. \ref{eq:sample_prior}
		\end{enumerate}
		\caption{Blocked Gibbs sampler for IMMC}\label{alg:blockedGS}
	}
\end{algorithm}

In the forward pass of the Baum-Welch algorithm, we have to compute the state-probability at each time-step $t$ conditioned on the hidden state of the previous time-step $z_{t-1}$, the state transition indicator $\omega_{t-1}$, and the backward probabilities $\bm{m_{t+1, t}}$. Therefore, we first have to compute
\begin{equation}
	\omega_t \sim \mbox{Ber}\left( \frac{\widetilde{L}^{\mathsf{intra}}_{t}}{\sum_{j=1}\widetilde{L}^{\mathsf{inter}}_{t, j} + \widetilde{L}^{\mathsf{intra}}_{t}} \right),
\end{equation}
where 
\begin{equation}
	\begin{aligned}
		\widetilde{L}^{\mathsf{intra}}_{t} &\triangleq p\left( y_t | \mathbf{\theta}, \omega_t=0, z_t=z_{t-1}, p_t \right) \cdot m_{t+1,t}(z_{t-1}) \cdot \kappa\\
		\widetilde{L}^{\mathsf{inter}}_{t, j} &\triangleq p\left( y_t | \mathbf{\theta}, \omega_t=1, z_{t-1}, p_t \right) \cdot m_{t+1,t}(j).
	\end{aligned}
\end{equation}
If $y_t$ or $p_t$ is the boundary state, $\omega_t$ is set to $0$ or $1$, respectively.\\
Given a realization of $\omega_t$, we can compute the probability distribution over the latent states at time-step $t$ by
\begin{equation}
    p(z_t|\bullet)\hspace{-0.1cm} \propto \hspace{-0.1cm}\left\{
    \begin{array}{l l}
        \hspace{-0.15cm}\mathbb{I}(z_t = z_{t-1}) & \mbox{if } y_t = B \mbox{ or } \omega_{t-1}=0\\
        \hspace{-0.15cm}\rho_t \cdot m_{t+1,t}(z_t) & \mbox{if } p_t = B \mbox{ or } \omega_{t-1}=1
    \end{array}\right.  
\end{equation}
with 
\begin{equation}
	\begin{aligned}
		p(z_t|\bullet) &\triangleq p(z_t|\omega_{t-1}, z_{t-1}, \bm{p}, \bm{y}, \bm{\pi}, \bm{\psi}, \bm{\theta}, \bm{m})\\
		\rho_t &\triangleq p(y_t|\bm{\theta}, z_t, p_t) \cdot p(z_t|\bm{\beta}).
	\end{aligned}
\end{equation}


Finally, the assignments are sampled from the computed probability distribution for $z_t$,
\begin{equation} \label{eq:sample_hidden}
    z_t \sim \mbox{Mu}\left(\frac{\sum_{i=1}^L p(z_t=i|\bullet) \mathbbm{I}(z_t = i)}{\sum_{i=1}^L p(z_t=i|\bullet)}\right).
\end{equation}

\input{combo}

During the sampling process, the auxiliary variables keep track of the sufficient statistics to update the prior distributions afterwards. Given a realization of $\bm{z}$, the prior distributions of the parameters are updated accordingly,
\begin{equation}
	\begin{aligned} \label{eq:sample_prior}
	    \bm{\beta}  &\sim \mbox{Dir}(\gamma / L + d_1, \dots, \gamma / L + d_L)\\
	    \bm{\pi_i}  &\sim \mbox{Dir}(\alpha\bm{\beta} + n_{i,\cdot} + \kappa\delta_i)\\
	    \bm{\psi}_i &\sim \mbox{Dir}(\sigma / K + G_{i, \cdot, 1}, \dots, \sigma / K + G_{i, \cdot, K})\\ \bm{\theta}_{i,k} &\sim \mbox{Dir}(\lambda / K + G_{i,k,1}, \dots, \lambda / K + G_{i,k,K}),
	\end{aligned}
\end{equation}

where $G_{i, \cdot, j}$ denotes the count of element $j \in \Sigma$ in super state $i$, $G_{i, \cdot, j} = \sum_{k = 1}^K G_{i, k, j}$.\\
Algorithm \ref{alg:blockedGS} summarizes the entire blocked Gibbs sampler.

%% file: combo.tex
    \begin{figure}[h]
		\scalebox{.95}{
            \centering
            \input{sc3}
		}
        \captionof{figure}{Generative processes of the test case III.}\label{fig:sc3}
    \end{figure}
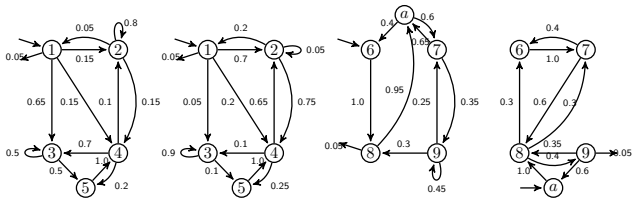

%% file: sc3.tex
  \scalebox{.65}{
  \begin{tikzpicture}[->,>=stealth',shorten >=1pt,auto,node distance=2cm,
    thick,main node/.style={circle,fill=blue!20,draw,font=\sffamily\Small\bfseries}, x=1.0cm,y=1.8cm, baseline=(current bounding box.north)]

    \node[latent, minimum size=4mm]                               (sc3_1a) {$1$};
    \node[latent, right=of sc3_1a, minimum size=4mm]              (sc3_2a) {$2$};
    \node[latent, below=of sc3_1a, minimum size=4mm]              (sc3_3a) {$3$};
    \node[latent, right=of sc3_3a, minimum size=4mm]              (sc3_4a) {$4$};
    \node[latent, right=of sc3_3a, minimum size=4mm, xshift=-0.7cm, yshift=-0.75cm] (sc3_5a) {$5$};
    \node[const,  left=of  sc3_1a, minimum size=4mm, xshift=0.5cm, yshift=0.3cm]    (sc3_1sa) {};
    \node[const,  left=of  sc3_1a, minimum size=4mm, xshift=0.5cm, yshift=-0.3cm]   (sc3_1ea) {};

    \path[every node/.style={font=\sffamily\tiny}]
    (sc3_1sa)edge              node [auto]  {}     (sc3_1a)
    (sc3_1a) edge              node [below] {0.15} (sc3_2a)
             edge              node [left]  {0.65} (sc3_3a)
             edge              node [left]  {0.15} (sc3_4a)
             edge              node [left]  {0.05} (sc3_1ea)
    (sc3_2a) edge [bend right] node [above] {0.05} (sc3_1a)
             edge [loop above] node [right] {0.8}  (sc3_2a)
             edge [bend left]  node [auto]  {0.15} (sc3_4a)
    (sc3_3a) edge [loop left]  node [left]  {0.5}  (sc3_3a)
             edge              node [left]  {0.5}  (sc3_5a)
    (sc3_4a) edge              node [auto]  {0.1}  (sc3_2a)
             edge              node [above] {0.7}  (sc3_3a)
             edge [bend left]  node [auto]  {0.2}  (sc3_5a)
    (sc3_5a) edge              node [above] {1.0}  (sc3_4a);

    \node[latent, right=of sc3_2a, xshift=0.5cm, minimum size=4mm](sc3_1b) {$1$};
    \node[latent, right=of sc3_1b, minimum size=4mm]              (sc3_2b) {$2$};
    \node[latent, below=of sc3_1b, minimum size=4mm]              (sc3_3b) {$3$};
    \node[latent, right=of sc3_3b, minimum size=4mm]              (sc3_4b) {$4$};
    \node[latent, right=of sc3_3b, minimum size=4mm, xshift=-0.7cm, yshift=-0.75cm] (sc3_5b) {$5$};
    \node[const,  left=of  sc3_1b, minimum size=4mm, xshift=0.5cm, yshift=0.3cm]    (sc3_1sb) {};
    \node[const,  left=of  sc3_1b, minimum size=4mm, xshift=0.5cm, yshift=-0.3cm]   (sc3_1eb) {};

    \path[every node/.style={font=\sffamily\tiny}]
    (sc3_1sb)edge              node [auto]  {}     (sc3_1b)
    (sc3_1b) edge              node [below] {0.7}  (sc3_2b)
             edge              node [left]  {0.05} (sc3_3b)
             edge              node [left]  {0.2}  (sc3_4b)
             edge              node [left]  {0.05} (sc3_1eb)
    (sc3_2b) edge [bend right] node [above] {0.2}  (sc3_1b)
             edge [loop right] node [right] {0.05} (sc3_2b)
             edge [bend left]  node [auto]  {0.75} (sc3_4b)
    (sc3_3b) edge [loop left]  node [left]  {0.9}  (sc3_3b)
             edge              node [left]  {0.1}  (sc3_5b)
    (sc3_4b) edge              node [auto]  {0.65} (sc3_2b)
             edge              node [above] {0.1}  (sc3_3b)
             edge [bend left]  node [auto]  {0.25} (sc3_5b)
    (sc3_5b) edge              node [above] {1.0}  (sc3_4b);

    \node[latent, right=of sc3_2b, xshift=0.65cm, minimum size=4mm](sc3_6a) {$6$};
    \node[latent, right=of sc3_6a, minimum size=4mm]              (sc3_7a) {$7$};
    \node[latent, below=of sc3_6a, minimum size=4mm]              (sc3_8a) {$8$};
    \node[latent, right=of sc3_8a, minimum size=4mm]              (sc3_9a) {$9$};
    \node[latent, right=of sc3_6a, minimum size=4mm, xshift=-0.7cm, yshift=0.75cm] (sc3_aa) {$a$};
    \node[const,  left=of  sc3_6a, minimum size=4mm, xshift=0.5cm, yshift=0.3cm]   (sc3_2sa) {};
    \node[const,  left=of  sc3_8a, minimum size=4mm, xshift=0.5cm, yshift=0.3cm]   (sc3_2ea) {};

    \path[every node/.style={font=\sffamily\tiny}]
    (sc3_2sa)edge              node [auto]  {}     (sc3_6a)
    (sc3_6a) edge              node [left]  {1.0}  (sc3_8a)
    (sc3_7a) edge [bend left]  node [right] {0.35} (sc3_9a)
             edge              node [below] {0.65} (sc3_aa)
    (sc3_8a) edge [bend right] node [left]  {0.95} (sc3_aa)
             edge              node [left]  {0.05} (sc3_2ea)
    (sc3_9a) edge              node [auto]  {0.25} (sc3_7a)
             edge              node [above] {0.3}  (sc3_8a)
             edge [loop below] node [auto]  {0.45} (sc3_9a)
    (sc3_aa) edge              node [above] {0.4}  (sc3_6a)
             edge [bend left]  node [above] {0.6}  (sc3_7a);

    \node[latent, right=of sc3_7a, xshift=0.35cm, minimum size=4mm] (sc3_6b) {$6$};
    \node[latent, right=of sc3_6b, minimum size=4mm]                (sc3_7b) {$7$};
    \node[latent, below=of sc3_6b, minimum size=4mm]                (sc3_8b) {$8$};
    \node[latent, right=of sc3_8b, minimum size=4mm]                (sc3_9b) {$9$};
    \node[latent, right=of sc3_8b, minimum size=4mm, xshift=-0.7cm, yshift=-0.75cm] (sc3_ab) {$a$};
    \node[const,  left=of  sc3_ab, minimum size=4mm, xshift=0.5cm]  (sc3_2sb) {};
    \node[const,  right=of  sc3_9b, minimum size=4mm, xshift=-0.5cm](sc3_2eb) {};

    \path[every node/.style={font=\sffamily\tiny}]
    (sc3_2sb)edge              node [auto]  {}     (sc3_ab)
    (sc3_6b) edge              node [below] {1.0}  (sc3_7b)
    (sc3_7b) edge [bend right] node [above] {0.4}  (sc3_6b)
             edge              node [left]  {0.6}  (sc3_8b)
    (sc3_8b) edge              node [left]  {0.3}  (sc3_6b)
             edge [bend right] node [above] {0.3}  (sc3_7b)
             edge [bend right] node [auto]  {0.4}  (sc3_9b)
    (sc3_9b) edge              node [above] {0.35} (sc3_8b)
             edge              node [right] {0.6}  (sc3_ab)
             edge              node [right] {0.05} (sc3_2eb)
    (sc3_ab) edge              node [left]  {1.0}  (sc3_8b);

  \end{tikzpicture}}


%% file: experiments.tex
\section{Experiments}
\label{sec:exp}

\newcommand*{\thead}[1]{\multicolumn{1}{c}{\bfseries #1}}

\subsection{Synthetic Data}
We evaluate the segmentation performance of our model to understand its effectiveness and test it for extreme cases. \footnote{The source code is available at: $<$\emph{anonymized}$>$.}
Therefore, we apply our model to three synthetic test cases which consist of generating processes, each emulating a different super state. 

These test cases differ primarily in their level of difficulty of identifying the processes (super states) correctly, with test case I being the least difficult and test case III the most difficult. Specifically, test case I is comprised of processes with no overlapping state spaces, meaning each state belongs to exactly one super state, while test case III features processes of completely overlapping state spaces that only differ in their inner dynamics, with test case II presenting both scenarios.
Given the synthetic nature of our test we can accurately evaluate the segmentation performance of our approach.


For each test-case, we generate three synthetic data sets to assess the performance of the algorithm with different amounts of data.
Figure \ref{fig:sc3} shows the generative processes of test case III where states are indexed by hexadecimal numbers. Realizations of these processes are sampled as segments and combined into sequences. The data sets are comprised of a set of these sequences which sum up to a total amount of $2,500$, $25,000$, and $250,000$ observations, respectively.

\begin{figure*}
\centering
  \vskip-\abovecaptionskip
  \captionof{table}[t]{Error rates for the artificial segmentation tasks; on average, each data set consists of $30,000$ data points.}\label{tab:prec_acc}
  \begin{tabular}{l|r|rrr|rrr}
    \toprule
     & \multicolumn{1}{c|}{\textbf{FMMC}} & \multicolumn{3}{c|}{\textbf{HHMM}} & \multicolumn{3}{c}{\textbf{IMMC}}\\
	&  \multicolumn{1}{c|}{(mid)} & \multicolumn{1}{c}{(small)} & \multicolumn{1}{c}{(mid)} & \multicolumn{1}{c|}{(large)} & \multicolumn{1}{c}{(small)} & \multicolumn{1}{c}{(mid)} & \multicolumn{1}{c}{(large)} \\
	\cline{1-8}
    Test-case   I   & $3.64\%^*$ & \cellcolor[HTML]{C0C0C0} $0.00\%$ & \cellcolor[HTML]{C0C0C0} $0.00\%$ & $14.38\%$ & $0.01\%$ & $0.01\%$ & \cellcolor[HTML]{C0C0C0} $0.01\%$ \\
    Test-case  II  & \cellcolor[HTML]{C0C0C0} $2.38\%^*$ & $8.99\%$ & $14.28\%$ & $11.13\%$ & \cellcolor[HTML]{C0C0C0} $3.35\%$ & $3.17\%$ & \cellcolor[HTML]{C0C0C0} $3.21\%$ \\
    Test-case III & $15.66\%^*$ & $11.00\%$ & $14.74\%$ & $15.84\%$ & \cellcolor[HTML]{C0C0C0} $0.01\%$ & \cellcolor[HTML]{C0C0C0} $0.01\%$ & \cellcolor[HTML]{C0C0C0} $0.01\%$ \\
    \bottomrule
  \end{tabular}%
  \caption*{* Segmentation-information provided; evaluated on the mid-sized data set.}
\end{figure*}

\input{sc2}

\paragraph*{Segmentation performance.}
In order to assess the segmentation performance of our algorithm, we evaluate the precision of identifying the processes and observation assignments of the synthetic data sets. As a baseline, we compare our results against those of the HHMM \citep{wakabayashi2012forward}. HHMMs are a logical choice as they fulfill our requirements for the segmentation (i-iii). Additionally, we also consider FMMC \citep{cadez2000visualization} as a baseline due to its close proximity in concept to IMMC. This approach represents a parametric interpretation of mixture models of Markov chains. Due to its lack of flexibility, it is unable to actually segment sequences, but it rather clusters them. Thus, we provide information on segment boundaries to this baseline. We ran the algorithm ten times with varying cluster initializations for each recorded result of our algorithm and the HHMM, and only selected the best result of FMMC for comparison. For HHMM we performed a grid-search to determine the optimal size of the state space. Each HHMM model was trained in $1000$ iterations. It is to note, that we achieved the best results with a larger state space than the actual one. 
For IMMC we report on results based on $250$ iterations with a burn-in phase of $250$ iterations.
All results are reported as the average of $10$ recorded runs.

Table \ref{tab:prec_acc} depicts error rates for the segmentation task. Even though FMMC has additional information, our approach outperforms it in both test case I and III. It seems that the provided segment boundary information is even more vital in test case II than in the other ones, as the first super state (see Figure \ref{fig:sc2}) consists of two loosely connected sub-graphs.\\
IMMC performs equally well over all data sets of any specific test case. Its performance seems unaffected by the amount of data provided.

For test case I, whose purpose is to evaluate the basic segmentation ability, the HHMM achieves a perfect result, closely followed by our algorithm. While scoring the perfect result on both the small- and mid-sized data set, the HHMM struggles with the large data set where its performance drops drastically.

Test case II demands a segmentation based on not only the distribution over sub-state spaces, but also on the dynamics within a super state. While the performance of our algorithm only slightly decreases (accuracy of $3.35\%$/$3.17\%$/$3.21\%$), the HHMM struggles with the more detailed segmentation task, achieving an accuracy of $8.99\%$/$14.28\%$/$11.13\%$, respectively. A reason that our algorithm performs worse than the FMMC (also IMMCs poorest performance over all data sets) seems to be the sloppy designed generating process (Figure \ref{fig:sc2} (first from left)) which contains two loosely connected sub-graphs.

Test case III heavily focuses on segmentation based on the inherent dynamics of the super states. Again, our algorithm outperforms the HHMM, scoring almost perfect accuracy results. In general, the results confirm the ability of HHMMs to perform basic segmentation tasks. Nonetheless, the algorithm seems to struggle with an increased complexity in the data induced by increasing the amount of data, as well as with segmentation tasks demanding distinction by both, sub-state space distributions and dynamics within super states.

Our algorithm performs consistently well over all test cases and data set sizes. On the data sets of test case III, it significantly outperforms both, the HHMM and the FMMC. Of further note is the insensibility to the set of hyper-parameters in our model, meaning rule-of-thumb adjustments should suffice. For the evaluations we used the same hyper-parameter values for all test cases over all data sets.

\subsection{User Navigation on Facebook}
The data set for the next evaluation contains user navigation data from Facebook \citep{puscher}. For each user, the invoked pages are recorded and grouped into sessions. Examples for such invoked pages are \emph{'Login', 'Newsfeed', 'Load more news', 'Like', etc}. The dataset contains $152$ unique invoked pages, $49,479$ sessions of $2,749$ users, and $8,197,308$ observations.
Every session is interpreted as a sequence of observations.

\input{cl_example}

\paragraph*{Prediction performance.}
To show the applicability of IMMC in the context of real-world applications, we measure its prediction performance on the Facebook data set. Therefore, we split the Facebook data into a training- and a test set using $90\%$ of the data for training and $10\%$ for testing. Furthermore, we cut each sequence of the test set at a randomly sampled position $c$ and use the sub-sequence $y^{(s)}_{1:c}$ as input to the model. The ground-truth for the prediction is the observation at position $c+1$. This situation simulates the prediction of future observations in a sequence given only past and present observations.
For the prediction process, we learn a model of the underlying super states given the training set. Conditioned on the observed sub-sequence, $y^{(s)}_{1:c}$, we compute the MAP estimate of the next state of the sequence based on the likelihoods of all super states and the transition probabilities within each super state from the most recent state to all possible future states.\\
The evaluation is performed three times: once on the entire data set, once on $10\%$ of the data, and once on $1\%$ of it. To show the influence of the detailed partition our algorithm applies to the data, we compare it to FMMC and to global Markov models of different orders. For FMMC we performed a grid-search to find the optimal size of the state space, i.e. the optimal number of MCs.

Whereas the MMs (order $\leq 9$) achieved an accuracy of $\approx 1.0\%$ on the entire data set, FMMC predicts $9.84\%$ of the cases correctly. Our algorithm, representing a more flexible version of FMMC, outperforms the other algorithms significantly. It results in a model with $61.41\%$ prediction-accuracy on the entire data set and slightly decreased performances on the smaller data sets, i.e. $57.81\%$ and $54.34\%$ on $10\%$ and $1\%$ of the data, respectively. 

\paragraph*{Runtime.}
Another important advantage to note is the computational efficiency of our algorithm compared to HHMMs. When evaluating the prediction performance of both algorithms on the Facebook data set, we noticed the significantly higher runtime of the HHMM. The evaluations were performed on a PC with an Intel Core i5-6600K CPU @ 3.50GHz and 4 cores, 32GB of RAM, a SSD, and a $64$-bit system. While both algorithm had almost identical computation times for a single iteration of less than a second on the mid-sized synthetic data sets, we terminated the computation of an iteration of the HHMM on the entire Facebook data set after several days. IMMC computed an iteration on the same data set in $\approx 5,379s$.
Computation times on $1\%$ of the Facebook data were $\approx2,985s$ and $\approx22s$ for the HHMM and our algorithm, respectively.\\
Given the unreliability of HHMMs with a low number of iterations and impractically high runtime of a sufficient number of iterations ($1\%$ of the Facebook data @ 1000 iterations: $> 30$ days), we were unable to even match the prediction performance of simple MMs.\\
In addition to the fast computation times of our algorithm we also obtain a fast convergence rate. 
On the synthetic data set the results were largely converged after only $40$ iterations (+$40$ burn-in iterations) while the learning process could be terminated on the Facebook data set after only $20$ iterations (+$20$ burn-in iterations).
The code of the HHMM was provided by \citet{wakabayashi2012forward}.

\paragraph*{Interpretability.}
Finally, we demonstrate how the model can be applied for information extraction tasks. This is especially useful for tasks that come with no or only little prior knowledge. Being a nonparametric model that adjusts its complexity to the data, our approach is a promising candidate for such tasks. Additionally, representing clusters by Markov models makes it easy to interpret the resulting segments. Figure \ref{fig:cl_exmpl} depicts three frequently observed behavioral patterns of users on Facebook.  \emph{(1)} shows  a user checking for updates on the \emph{newsfeed} or waiting for \emph{new messages}. The user \emph{activates} the Facebook tab and without doing any additional activity \emph{deactivates} it shortly after. \emph{(2)}  represents users communicating with each other. \emph{(3)} shows users who are interested in updates of their friends. After \emph{activating} the Facebook tab, scrolling the \emph{newsfeed} and visiting specific \emph{newsfeed entries}, users \emph{deactivate} the tab again. These types of segments represent user behavior focused on specific tasks. Our results give a  detailed insight in how users interact on Facebook.

%% file: sc2.tex
    \begin{figure*}
  \centering
  \scalebox{.7}{
  \begin{tikzpicture}[->,>=stealth',shorten >=1pt,auto,node distance=3cm,
    thick,main node/.style={circle,fill=blue!20,draw,font=\sffamily\Small\bfseries}]

    \node[const] (s1a) {};
    \node[latent, above=of s1a, minimum size=4mm, xshift=0.8cm, yshift=-0.8cm] (11) {$1$};
    \node[latent, right=of 11,  minimum size=4mm]             (12) {$2$};
    \node[const,  right=of 12,  xshift=-0.4cm]                (s1b)   {};
    \node[latent, above=of 11,  minimum size=4mm]             (13) {$3$};
    \node[latent, right=of 13,  minimum size=4mm]             (14) {$4$};
    \node[latent, right=of 14,  minimum size=4mm]             (15) {$5$};
    \node[const,  below=of 15,  minimum size=4mm, yshift=0.4cm] (e1) {};

    \path[every node/.style={font=\sffamily\tiny}]
    (s1a) edge              node [auto]            {0.3}    (11)
    (s1b) edge              node [auto]            {0.7}    (12)
    (11)  edge              node [auto]            {0.9}    (12)
          edge              node [auto]            {0.1}    (13)
    (12)  edge [bend left]  node [auto]            {0.95}   (11)
          edge              node [left]            {0.05}   (13)
    (13)  edge [loop left]  node [auto]            {0.55}   (13)
          edge [bend left]  node [auto, near end]  {0.05}   (12)
          edge              node [below, near end] {0.4}    (14)
    (14)  edge [bend right] node [above]           {0.05}   (13)
          edge              node [below]           {0.95}   (15)
    (15)  edge [bend right] node [above]           {0.4}    (14)
          edge [loop right] node [auto]            {0.4}    (15)
          edge              node [auto]            {0.2}    (e1);

    \tiny
    \node[const,  right=of s1a,  xshift=3.6cm]   (s3)    {};
    \node[latent, above=of s3, minimum size=4mm, xshift=0.8cm, yshift=-0.8cm] (39) {$9$};
    \node[latent, right=of 39,  minimum size=4mm]             (310) {$a$};
    \node[latent, above=of 39,  minimum size=4mm]             (311) {$b$};
    \node[latent, right=of 311, minimum size=4mm]             (312) {$c$};
    \node[const,  right=of 312, minimum size=4mm, xshift=-0.5cm, yshift=-0.6cm] (e3) {};

    \path[every node/.style={font=\sffamily\tiny}]
    (s3)  edge              node [auto]  {1.0}    ( 39)
    (39)  edge              node [auto]  {0.6}    (310)
          edge              node [right] {0.4}    (311)
    (310) edge [bend left]  node [auto]  {1.0}    (39)
    (311) edge [bend right] node [left]  {0.2}    (39)
          edge              node [auto]  {0.55}   (310)
          edge [loop above] node [auto]  {0.1}    (311)
          edge              node [auto]  {0.15}   (312)
    (312) edge [loop above] node [auto]  {0.7}    (312)
          edge              node [auto]  {0.3}    (e3);

    \tiny
    \node[const,  right=of 310, xshift=0.8cm]   (s5)    {};
    \node[latent, right=of s5,  minimum size=4mm, xshift=-0.4cm] (515) {$f$};
    \node[latent, above=of 515, minimum size=4mm]  (56) {$6$};
    \node[latent, left=of 56,   minimum size=4mm] (514) {$e$};
    \node[latent, right=of 515, minimum size=4mm]  (54) {$4$};
    \node[latent, right=of 56,  minimum size=4mm] (512) {$c$};
    \node[const,  below=of 514, minimum size=4mm, yshift=0.4cm] (e5) {};

    \path[every node/.style={font=\sffamily\tiny}]
    (s5)  edge              node [auto]  {1.0}    (515)
    (515) edge              node [auto]  {0.85}   ( 54)
          edge [loop below] node [left]  {0.15}   (515)
    (54)  edge              node [auto]  {1.0}    (512)
    (512) edge [bend left]  node [auto]  {0.1}    ( 54)
          edge              node [auto]  {0.7}    ( 56)
          edge [loop above] node [auto]  {0.2}    (512)
    (56)  edge              node [auto]  {0.75}   (515)
          edge [bend right] node [above] {0.25}   (514)
    (514) edge              node [below] {0.15}   ( 56)
          edge [loop above] node [auto]  {0.25}   (514)
          edge              node [auto]  {0.6}    ( e5);

    \node[latent, right=of 54, minimum size=4mm, xshift=0.6cm] (26) {$6$};
    \node[latent, above=of 26, minimum size=4mm, xshift=-0.8cm](27) {$7$};
    \node[latent, above=of 26, minimum size=4mm, xshift=0.8cm] (28) {$8$};
    \node[const,  below=of 26, yshift=0.5cm]                   (s2)    {};
    \node[const,  above=of 28, minimum size=4mm, xshift=0.8cm, yshift=-0.6cm] (e2) {};

    \path[every node/.style={font=\sffamily\tiny}]
    (s2) edge              node [auto]  {1.0} (26)
    (26) edge              node [auto]  {0.6} (27)
         edge              node [left]  {0.1} (28)
         edge [loop left]  node         {0.3} (26)
    (27) edge              node [below] {1.0} (28)
    (28) edge [bend left]  node [right] {0.2} (26)
         edge [bend right] node [above] {0.6} (27)
         edge              node [auto]  {0.2} (e2);

    \tiny
    \node[latent, right=of 28, minimum size=4mm, xshift=0.2cm] (41)  {$1$};
    \node[latent, below=of 41, minimum size=4mm]               (48)  {$8$};
    \node[latent, right=of 41, minimum size=4mm]               (410) {$a$};
    \node[latent, right=of 48, minimum size=4mm]               (412) {$c$};
    \node[latent, right=of 48, minimum size=4mm, xshift=-0.7cm, yshift=-0.75cm](413) {$d$};
    \node[const,  left=of 41, xshift=0.5cm]                    (s4b) {};
    \node[const,  below=of s4b, yshift=-0.36cm]                                (s4a) {};
    \node[const,  left=of 413, minimum size=4mm, xshift=0.5cm] (e4)  {};

    \path[every node/.style={font=\sffamily\tiny}]
    (s4a) edge              node [auto]  {0.6}    ( 48)
    (s4b) edge              node [auto]  {0.4}    ( 41)
    (48)  edge              node [auto]  {0.95}   (410)
          edge              node [left]  {0.05}   (413)
    (41)  edge              node [below] {0.99}   (410)
          edge              node [left]  {0.01}   ( 48)
    (410) edge [bend right] node [above] {0.2}    ( 41)
          edge [loop above] node [auto]  {0.3}    (410)
          edge [bend left]  node [auto]  {0.5}    (412)
    (412) edge              node [above] {0.65}   ( 48)
          edge              node [auto]  {0.35}   (410)
    (413) edge              node [right] {0.1}    (412)
          edge [loop right] node [auto]  {0.1}    (413)
          edge              node [auto]  {0.8}    ( e4);

    \tiny
    \node[latent, right=of 412, minimum size=4mm, xshift=0.2cm] (615) {$f$};
    \node[latent, above=of 615, minimum size=4mm]  (63) {$3$};
    \node[latent, right=of 615, minimum size=4mm]  (68) {$8$};
    \node[latent, right=of 63,  minimum size=4mm]  (64) {$4$};
    \node[latent, right=of 68,  minimum size=4mm]  (69) {$9$};
    \node[latent, right=of 64,  minimum size=4mm]  (65) {$5$};
    \node[const,  below=of 68,  minimum size=4mm, yshift=0.5cm] (s6) {};
    \node[const,  above=of 64,  minimum size=4mm, yshift=-0.5cm, xshift=0.6cm] (e6) {};

    \path[every node/.style={font=\sffamily\tiny}]
    (s6)  edge              node [right] {1.0}    ( 68)
    (68)  edge              node [auto]  {0.3}    ( 69)
          edge              node [auto]  {0.2}    ( 63)
          edge [bend left]  node [auto]  {0.5}    (615)
    (69)  edge              node [auto]  {0.6}    ( 65)
          edge [loop below] node [auto]  {0.4}    ( 69)
    (65)  edge [loop above] node [auto]  {0.4}    ( 65)
          edge [bend right] node [above] {0.6}    ( 64)
    (64)  edge [bend right] node [above] {0.3}    ( 63)
          edge              node [below] {0.5}    ( 65)
          edge              node [auto, near start] {0.2} (e6)
    (615) edge              node [auto]  {1.0}    ( 68)
    (63)  edge [loop above] node [auto]  {0.4}    ( 63)
          edge              node [left]  {0.3}    (615)
          edge              node [below] {0.3}    ( 64);

  \end{tikzpicture}}

  \caption{Generative processes of test case II; states are indexed by hexadecimal numbers (1-f).}
  \label{fig:sc2}
\end{figure*}
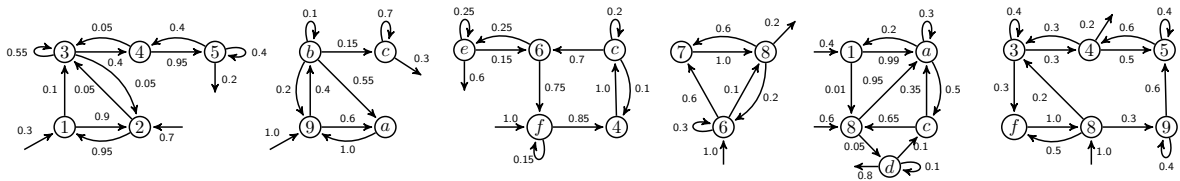

%% file: cl_example.tex
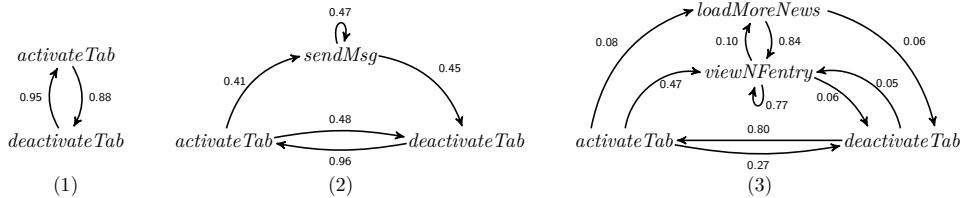
\begin{figure*}[t]
  \centering
  \scalebox{.75}{
  \begin{tikzpicture}[->,>=stealth',shorten >=1pt,auto,node distance=3cm,
    thick,main node/.style={circle,fill=blue!20,draw,font=\sffamily\Small\bfseries}]

    \node[const, minimum size=4mm]  (g321) { \emph{activateTab} };
    \node[const, minimum size=4mm, below=of g321, yshift=-0.1cm](g324) { \emph{deactivateTab} };

    \node[const, minimum size=4mm, right=of g324, xshift=-0.15cm] (aTab2) { \emph{activateTab} };
    \node[const, minimum size=4mm, right=of aTab2, xshift=-0.5cm, yshift=1.5cm] (send) { \emph{sendMsg} };
    \node[const, minimum size=4mm, right=of send, xshift=-0.5cm, yshift=-1.5cm] (dTab2) {\emph{deactivateTab}};

    \node[const, minimum size=4mm, right=of dTab2, xshift=-0.15cm] (aTab) { \emph{activateTab} };
    \node[const, minimum size=4mm, right=of aTab, xshift=-0.5cm, yshift=1.2cm] (vNFe) { \emph{viewNFentry} };
    \node[const, minimum size=4mm, above=of vNFe, yshift=-0.3cm]  (lmon) { \emph{loadMoreNews} };
    \node[const, minimum size=4mm, right=of vNFe, xshift=-0.5cm, yshift=-1.2cm] (dTab) {\emph{deactivateTab}};

    \node[const, below=of g324, yshift=0.6cm] (n1) {(1)};
    \node[const, right=of n1,   xshift=3.4cm] (n2) {(2)};
    \node[const, right=of n2,   xshift=5.95cm] (n3) {(3)};

    \path[every node/.style={font=\sffamily\tiny}]
    (vNFe)     edge [loop below]                node [right]{0.77} (vNFe)
               edge [bend left]                 node [left] {0.10} (lmon)
    (vNFe.355) edge [bend left=20]              node [left] {0.06} (dTab.160)
    (aTab)     edge [bend left=45]              node [right]{0.47} (vNFe.180)
    (aTab.355) edge [bend right=10]             node [below]{0.27} (dTab.185)
    (aTab.160) edge [bend left]                 node [auto] {0.08} (lmon.180)
    (lmon)     edge [bend left]                 node [right]{0.84} (vNFe)
    (lmon.0)   edge [bend left, shorten <=3pt]  node [auto] {0.06} (dTab.20)
    (dTab)     edge [bend right=45]             node [right]{0.05} (vNFe.0)
               edge                             node [above]{0.80} (aTab)

    (aTab2)     edge [bend left]    node [left]  {0.41} (send.180)
    (aTab2.3)   edge [bend left=10] node [auto]  {0.48} (dTab2.177)
    (send)      edge [loop above]   node [auto]  {0.47} (send)
    (send.0)    edge [bend left]    node [auto]  {0.45} (dTab2)
    (dTab2.183) edge [bend left=10] node [auto]  {0.96} (aTab2.357)

    (g321) edge [bend left]  node [auto] {0.88} (g324)
    (g324) edge [bend left]  node [auto] {0.95} (g321);

  \end{tikzpicture}}
  \caption{An examplary solution of the identified super states; exit nodes are omitted, their probability equals $1$ minus the sum of emission probabilities of a node.}
  \label{fig:cl_exmpl}
\end{figure*}

%% file: conclusion.tex
\section{Conclusion}
\label{sec:conc}

We presented a Bayesian nonparametric approach to perform a two-level analysis of the dynamics in categorical-valued time-series. By interpreting the two levels as the hidden states of an unbounded mixture model and the super states represented by Markov chains as its mixture components, our model showed significant improvements over related approaches when analyzing categorical-valued time-series. We obtained a natural state-duration model by augmenting both the hidden- and the observed layer of states. The hereby increased detail of the model allowed us to capture state durations based on the dynamics of the super states. Furthermore, by representing each super state by a Markov chain we obtained a model that yields easily interpretable low-level dynamics of the super states and achieves a highly accurate prediction rate. Thus, the model inherently is applicable to prediction- and information extraction tasks. 